\documentclass[lettersize,journal]{IEEEtran}
\usepackage{amsmath,amsfonts}

\usepackage{array}
\usepackage[caption=false,font=normalsize,labelfont=sf,textfont=sf]{subfig}
\usepackage{textcomp}
\usepackage{stfloats}
\usepackage{url}
\usepackage{verbatim}
\usepackage{graphicx}
\usepackage{cite}
\RequirePackage{multirow}

\usepackage{algpseudocode}
\usepackage{algorithm}
\usepackage{mathtools}
\usepackage{graphicx}
\usepackage{booktabs}
\usepackage{amsmath}
\usepackage{soul}
\usepackage[citecolor=blue, colorlinks]{hyperref}

\newcommand{\mypara}[1]{\noindent\textbf{#1.}~}
\usepackage[accsupp]{axessibility}

\begin{document}

\title{Modeling State Shifting via Local-Global Distillation for Event-Frame Gaze Tracking}

\author{Jiading Li, Zhiyu Zhu, Jinhui Hou, Junhui Hou, \textit{Senior Member, IEEE}, and Jinjian Wu,  \textit{Senior Member, IEEE}

\thanks{This work was supported in part by Hong Kong Research Grants Council under Grant 11218121, and in part by Hong
Kong Innovation and Technology Fund under Grant MHP/117/21. J. Li and Z. Zhu contributed equally to this work.}
\thanks{J. Li, Z. Zhu, J. Hou, and J. Hou are with the Department of Computer Science, City University of Hong Kong, Hong Kong SAR. Email: jiadingli3-c@my.cityu.edu.hk; zhiyuzhu2-c@my.cityu.edu.hk; jhhou3-c@my.cityu.edu.hk; jh.hou@cityu.edu.hk}
\thanks{J. Wu is with the School of Artificial Intelligence, Xidian
University, Xi’an 710071, China. Email: jinjian.wu@mail.xidian.edu.cn}
}

\markboth{Manuscript submitted IEEE TVCG}%
{Shell \MakeLowercase{\textit{et al.}}: A Sample Article Using IEEEtran.cls for IEEE Journals}

\maketitle

\begin{abstract}
\label{sec:abstract}
This paper tackles the problem of passive gaze estimation using both event and frame data. Considering the inherently different physiological structures, it is intractable to accurately estimate gaze purely based on a given state. Thus, we reformulate gaze estimation as the quantification of the state shifting from the current state to several prior registered anchor states. Specifically, we propose a two-stage learning-based gaze estimation framework that divides the whole gaze estimation process into a coarse-to-fine approach involving anchor state selection and final gaze location. Moreover, to improve the generalization ability, instead of learning a large gaze estimation network directly, we align a group of local experts with a student network, where a novel denoising distillation algorithm is introduced to utilize denoising diffusion techniques to iteratively remove inherent noise in event data. Extensive experiments demonstrate the effectiveness of the proposed method, which surpasses state-of-the-art methods by a large margin of 15$\%$. The code will be publicly available at \url{https://github.com/jdjdli/Denoise_distill_EF_gazetracker}.
\end{abstract}


\section{Introduction}
\label{sec:intro}

\IEEEPARstart
{E}{ye} gaze tracking represents a critical component in non-verbal communication, offering profound insights into an individual's intentions and emotional states by capturing visual information from the user's eyes. This technology can infer desires and needs by measuring the user's attention on specific objects, making it invaluable in fields like human-computer interaction \cite{Reiter2022look, choi2022kuiper, kim2022lattice}, virtual reality \cite{kassner2014pupil, ahn2021stickpie}, and intelligent transportation \cite{wee2018focusvr, ma2018combining}.

Over the recent decades, gaze estimation has witnessed an exponential increase in the development of diverse methodologies, which can be generally divided into three categories, i.e., 2D eye feature regression-based approaches~\cite{hennessey2003a,zhu2006nonlinear,wang2003eye,erroll2014eyetab}, 3D model-based eye movement reconstruction algorithms~\cite{lu2022neural}, and appearance-based gaze estimation techniques~\cite{rajeev2018light}. 
Although the first two categories excel in tracking the eye's position and movement, they typically rely on specialized hardware systems and light conditions.
Conversely, appearance-based methods only utilize face or eye images as input to train a mapping model between appearance and gaze, thereby determining the corresponding gaze based on new appearance data~\cite{baluja1993Non-Intrusive,Krafka2016eye,rajeev2018light,cheng2023dvgaze}.
However, appearance-based methods are sensitive to individual differences and head movements in unconstrained environments. Addressing these challenges often necessitates high-speed, high-resolution RGB and optical cameras to capture more detailed visual information, leading to significant costs and energy demands~\cite{Koushik2022a,Lystbaek2022Exploring}.

\begin{figure*}[t]
    \centering
    \resizebox{1\textwidth}{!}{\includegraphics[trim={0cm 10cm 0cm 2.2cm},clip]{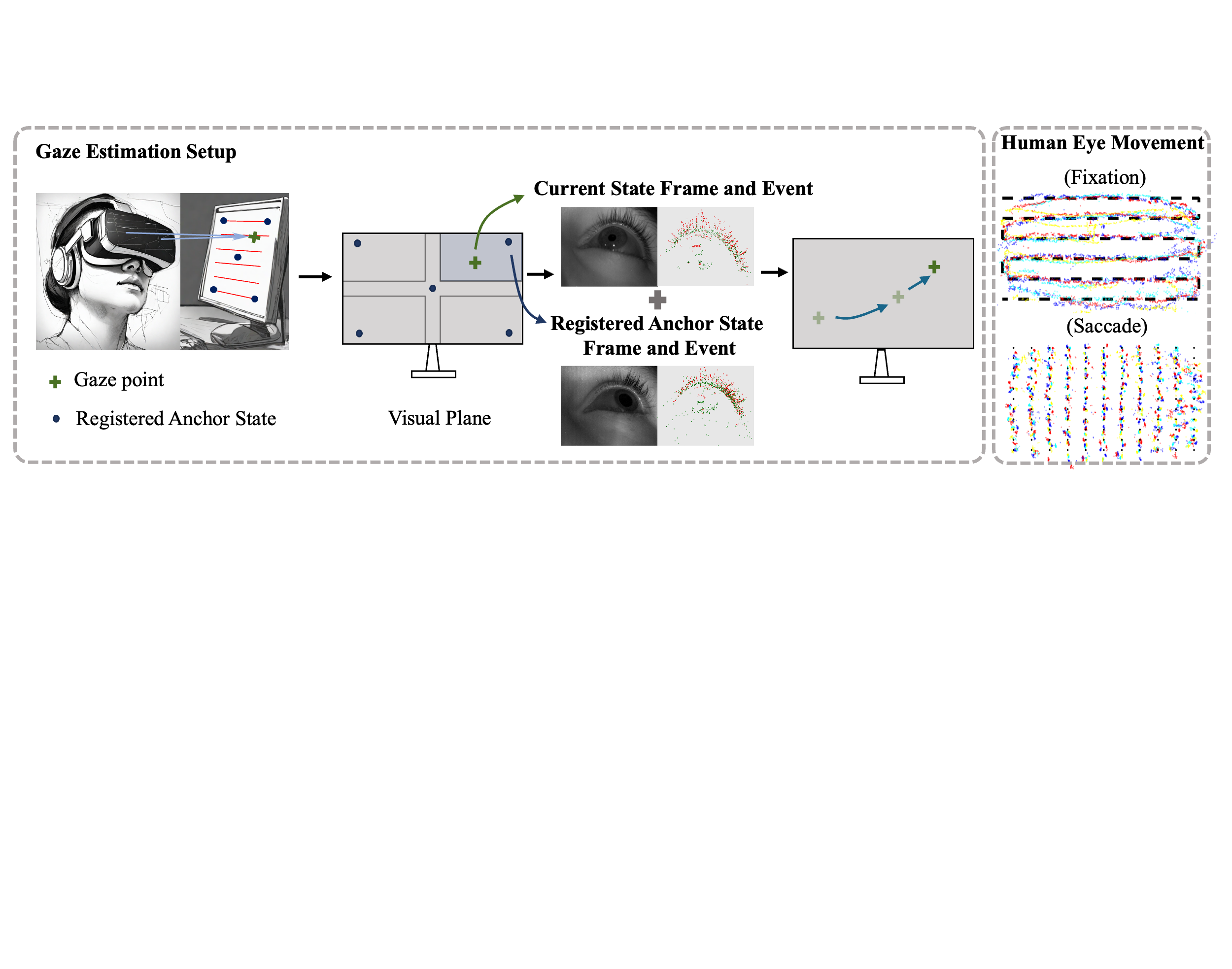}}
      \vspace{-0.4cm}
    \caption{\textbf{Left:} Overview of our gaze estimation setup: our framework emphasizes the modeling of gaze shifts from a registered anchor state to the currently acquired state captured during actual use. Our approach takes input in the form of a frame coupled with corresponding event data to infer the position of the directional gaze point as the output. \textbf{Right:} Beyond the confines of static frame-based gaze estimation, studying dynamic ocular movements constitutes an additional research trajectory within computer vision.}
    \label{fig:introduction}
\end{figure*}

Recently, event-based cameras have surged in popularity due to their exceptional temporal resolution, low latency, and expansive dynamic range, making them ideal for tracking fast eye movements like saccades with high fidelity~\cite{Angelopoulos2021event}. 
Additionally, the high dynamic range allows them to operate well under varying lighting conditions, which is crucial for accurate gaze estimation in real-world scenarios.
Despite the advantages, typical event-based sensors provide limited visual information, missing out on color, texture, and comprehensive contextual details that are readily available through conventional RGB imaging\cite{Gehrig2018Asynchronous,liu2016combined,Mokatren20233D,cheng2022gaze}. 
Therefore, the emerging field of cross-modal gaze estimation promises to enhance the robustness and accuracy of gaze tracking by capitalizing on the complementary strengths of both data streams~\cite{Leutenegger2015keyframebased,feng2022real,lei2023an,ansari2023person,zhao2023ev-eye}. However, most existing methods leverage only one modality to assist the other, failing to take full advantage of the information contained within both data types.

In this paper, we propose a novel gaze estimation transformer framework that revolutionizes gaze tracking by leveraging the complementary strengths of event cameras and traditional frame-based imaging, as shown in Fig.~\ref{fig:introduction}. Though synergizing the high temporal resolution of event data with the rich spatial information of frames via a local-global distillation process, our method achieves a new level of performance in gaze tracking. Technically, we reformulate the gaze estimation as the quantification of eye motion transitions from the current state to several prior registered anchor states. Based on this, we initially partition the entire gaze points region into several sub-regions and employ the vision transformers to pre-train a set of models on different sub-regions, yielding several \textbf{local expert networks} with relatively high accuracy for localized gaze estimation. Furthermore, we introduce a \textbf{local-global latent denoising distillation} method to distill knowledge from the set of local expert networks to a global student network to diminish the adverse effects of inherent noise from event data on student network performance.  Extensive experiments demonstrate the significant superiority of the proposed method over state-of-the-art methods.

In summary, the main contributions of this paper are three-fold:
\begin{itemize}
    \item we formulate the gaze estimation as an end-to-end prediction of state shifting from registered anchor state; 
    \item we distill multiple pre-trained local expert networks into a more robust student network to combat overfitting in gaze estimation; and 
    \item we propose a self-supervised latent denoising method to mitigate the adverse effects of noise from local expert networks to improve the performance of the student network.  
\end{itemize}

The remainder of this paper is organized as follows. Section~\ref{sec:related} briefly reviews related works in this field. Section~\ref{sec:method} presents the proposed method in detail, followed by extensive experiments and analysis in Section~\ref{sec:experiment}. Finally, Section~\ref{sec:conclusion} concludes this paper.

\begin{figure*}[t]
    \centering
    \resizebox{1\textwidth}{!}{\includegraphics[trim={0cm 1cm 0cm 1.3cm},clip]{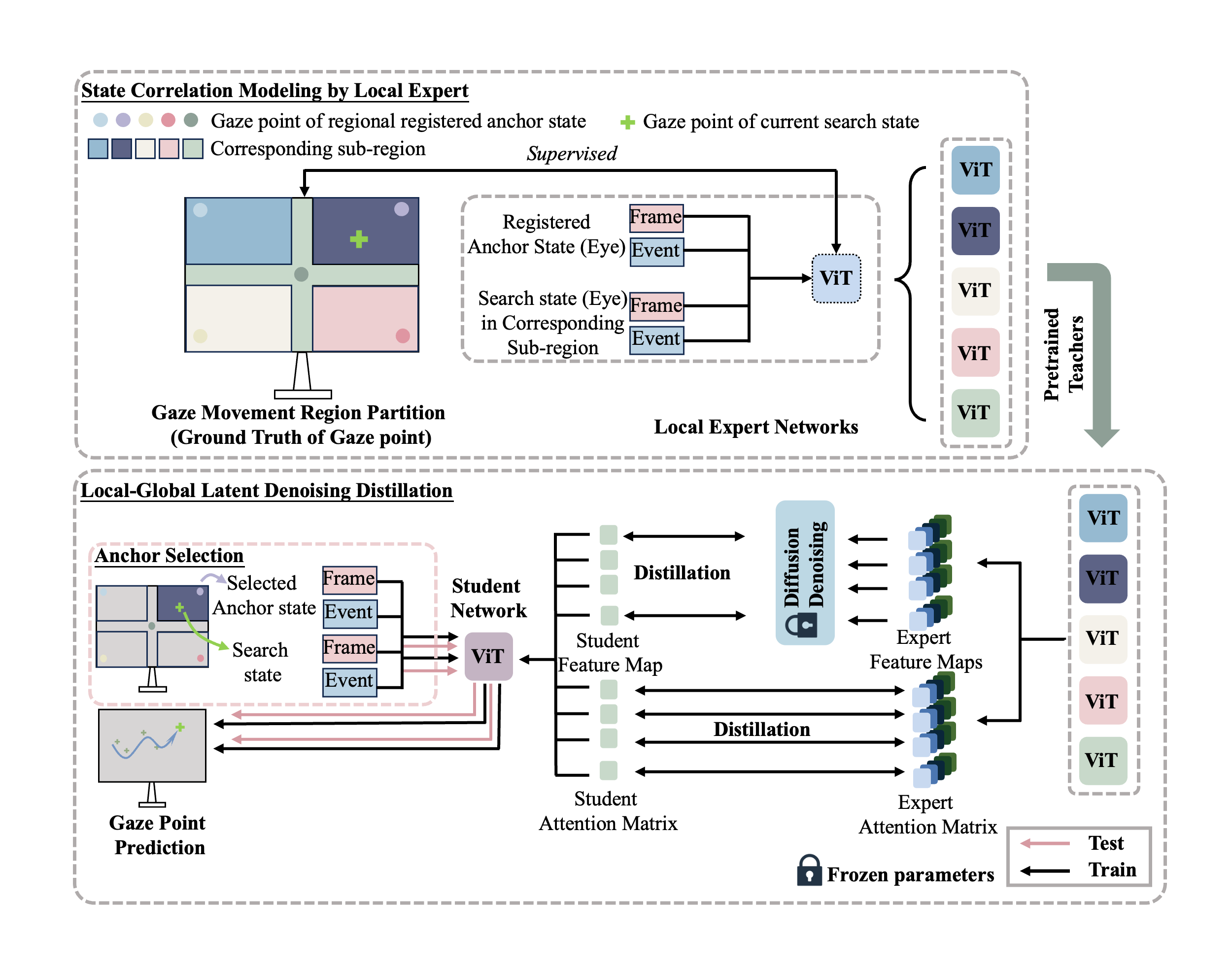}}
    \caption{Illustration of the workflow of the proposed framework, where \textbf{Black arrow} (resp. \textbf{Pink arrow}) represents the training (resp. testing) pipeline. \textbf{The First Stage (Sec.~\ref{Sec:Gaze_struct}): State Correlation Modeling by Local Expert.} We first partition the entire gaze points region into several sub-regions, wherein each region's data is trained to cultivate a local expert network. Each expert network is simultaneously fed with the anchor state and a search state and utilizes the transformers to explicitly model the correlation between the anchor and states. \textbf{The Second Stage (Sec.~\ref{Sec:Distillation}): Local-Global Latent Denoising Distillation.} A latent denoising knowledge distillation method is introduced to amalgamate the expertise of these several local expert networks into a singular, comprehensive student network. \textit{Note that the latent denoising and knowledge distillation are utilized in the training phase only (see details in Sec.~\ref{sec:experiment}). Anchor selection in the light pink box is illustrated in detail in Fig.~\ref{fig:anchorselection}.}
    }
    \label{fig:framework}
\end{figure*}

\section{Related work}
\label{sec:related}

In this section, we give a review of event-based vision, eye tracking, event-frame methods, and distillation networks.

\vspace{1em}
\noindent \textbf{Event-based Vision.} Neuromorphic event-based cameras, inspired by the Silicon Retina concept~\cite{mahowald1994an,carver1988a}, are key for fast vision tasks due to their quick response and low latency~\cite{Delbruck1993Silicon,Delbruck2010activity,Lichtsteiner2008a,posch2014Retinomorphic}. They support a wide range of applications including object recognition~\cite{Mitrokhin2018event,liu2016combined}, navigation~\cite{henri2017real}, pose estimation~\cite{Mueggler2017the}, 3D reconstruction~\cite{julien2018an}, SLAM~\cite{davison2007monoslam,Weikersdorfer2014event}, gesture tracking~\cite{lee2012touchless}, and object tracking~\cite{zhu2023cross,zhu2022learning,wang2023visevent,tang2022revisiting}. Initially, these cameras used event patterns to detect motion~\cite{delbruck2007fast,delbruck2013robotic,Litzenberger2006estimation,Litzenberger2006embedded} and track simple shapes~\cite{Lagorce2014Asynchronous,Conradt2009a}. Later improvements introduced event-driven algorithms~\cite{ni2012Asynchronous} and optimization techniques like gradient descent to refine tracking~\cite{ni2015visual}. Algorithms such as mean-shift and Monte Carlo~\cite{glover2017robust,Reverter2015an} have further enhanced tracking by adjusting to changes in the model. For instance, part-based models~\cite{Reverter2015an} have segmented subjects into parts, enabling quick and accurate tracking of facial or body movements~\cite{li2024e-gaze}. Chen \emph{et al.} ~\cite{chen20233et} introduced a sparse Change-Based ConvLSTM model for efficient event-based eye tracking. Stoffregen \emph{et al.} ~\cite{stoffregen2022eventbased} presented a novel method for high-frequency, low-power eye tracking using event cameras and a coded differential lighting scheme to enhance corneal glint detection while suppressing non-glint events. Wang~\emph{et al.}~\cite{wang2024longterm} introduced a unified single-stage transformer-based framework for efficient and accurate color-event object tracking. Ryan \emph{et al.} ~\cite{RYAN2021real} introduced a novel method for real-time face and eye tracking, as well as blink detection by using event cameras. It leverages a fully convolutional recurrent neural network architecture to enhance driver monitoring systems. However, the sparse nature of event data can be problematic in low-contrast settings, and using algorithms designed for dense data in sparse situations may still increase computational demands.

\vspace{1em}
\noindent \textbf{Eye Tracking.} Progressing from initial camera-based systems of eye tracking that monitored Purkinje images~\cite{young1975survey,Cornsweet1973accurate,crane1985generation,li2010eye,tian2000dual,wang2017real}, Morimoto \emph{et al.}~\cite{Morimoto2005eye} and Duchowski \emph{et al.}~\cite{Duchowski2017eye} provided detailed examinations of these pupil modeling and gaze estimation processes. Contemporary research focuses on deep learning to deduce gaze orientation from complex facial datasets obtained via standard webcams, directly mapping the visual characteristics of the eye to gaze coordinates, with Chen \emph{et al.}~\cite{chen2019appearance} enhancing accuracy through dilated convolution techniques. Advancing the field, Cheng \emph{et al.}~\cite{cheng2023dvgaze} integrated full face and eye region data for more accurate gaze inference and incorporates transformer models to exploit their superior handling of data dependencies. Nonetheless, these advanced models are best for full-face images and not ideal for eye-only cameras, requiring custom-designed networks for accurate data analysis. Also, their effectiveness is limited by the camera's frame rate, which can affect their real-time accuracy and performance.

\vspace{1em}
\noindent \textbf{Event-frame Methods.} Hybrid methods combine the detailed intensity data from a standard frame with the rapid detection of intensity changes from asynchronous event streams~\cite{feng2022real,lei2023an,ansari2023person,zhao2023ev-eye}. Wang~\emph{et al.}~\cite{wang2023visevent} introduced a cross-modality transformer algorithm for enhancing reliable object tracking by combining visible and event camera data, demonstrating improved performance in challenging scenarios. Feng \emph{et al.}~\cite{feng2022real} developed an event-driven eye segmentation algorithm that overcomes the limitations of standard frame rates, maintaining high accuracy despite a lower resolution. The auto-segmentation model was designed to combine with a previous gaze estimation model to improve estimation accuracy. In fact, it was not an end-to-end estimation model and the prediction latency was very significant. Angelopoulos \emph{et al.}~\cite{Angelopoulos2021event} enhanced the temporal resolution of gaze tracking by integrating event cameras close to the eyes. These cameras provided constant updates to the initial pupil location determined by traditional algorithms, allowing for precise adjustments at high temporal resolutions. However, the conventional baseline can significantly affect the performance of these methods~\cite{bao2022an,shin2023individual}. This approach resulted in gaze estimation accuracy varying within a wide range.

\vspace{1em}
\noindent \textbf{Distillation Networks.} Knowledge distillation~\cite{Hinton2015distilling} is intended for the transfer of learned features from a "teacher" model to an efficient "student" model. Wang~\emph{et al.}~\cite{Wang2024Event} presented a novel hierarchical knowledge distillation framework for high-speed, low-latency visual object tracking using event cameras. Lopez-Paz \emph{et al.}~\cite{Lopez2015unifying} expanded upon this concept by introducing privileged information, wherein the student model leverages insights from multiple teacher models, each accessing different data sources. Xiang \emph{et al.}~\cite{xiang2020learning} utilized the collective intelligence of several teacher models to tackle the long-tailed distribution problems. Guo \emph{et al.}~\cite{guo2020online} formulated a collaborative learning framework, similar to a congregation of local experts sharing their knowledge. Therefore, the essence of these expert networks' collective intelligence is distilled into a singular, more generalized student network~\cite{lucian2023jedi,shen2023expert,Sochopoulos2023deep}, with an enhanced level of accuracy, surpassing what could be achieved by an individual model trained on a uniform dataset. This kind of distillation has been proven to be suitable for the end-to-end gaze estimation task, with the detailed information provided in Sec.~\ref{sec:method}.

This study fuses conventional intensity frames with dynamic event camera data into a unified network architecture, leveraging pre-trained local expert models. It aims to utilize this integration to create an end-to-end, advanced, and accurate eye-tracking method, harnessing the unique benefits of both traditional frame modality and innovative event-based sensor input.

\section{Proposed Method}
\label{sec:method}

Gaze tracking aims to determine the gaze location, given the measured state. However, the high speed of eye movement and the subtle pattern of the eyeball make it hard to derive accurate predictions. Inspired by the high-temporal resolution and low-latency characteristics of event-based vision~\cite{Delbruck1993Silicon,Delbruck2010activity,Lichtsteiner2008a,posch2014Retinomorphic}, we propose to utilize frames together with event data for building an accurate near-eye gaze estimation pipeline, as illustrated in Fig.~\ref{fig:framework}.

\begin{figure*}[t]
    \centering
    \resizebox{1\textwidth}{!}{\includegraphics[trim={0.1cm 7cm 0cm 5.3cm},clip]{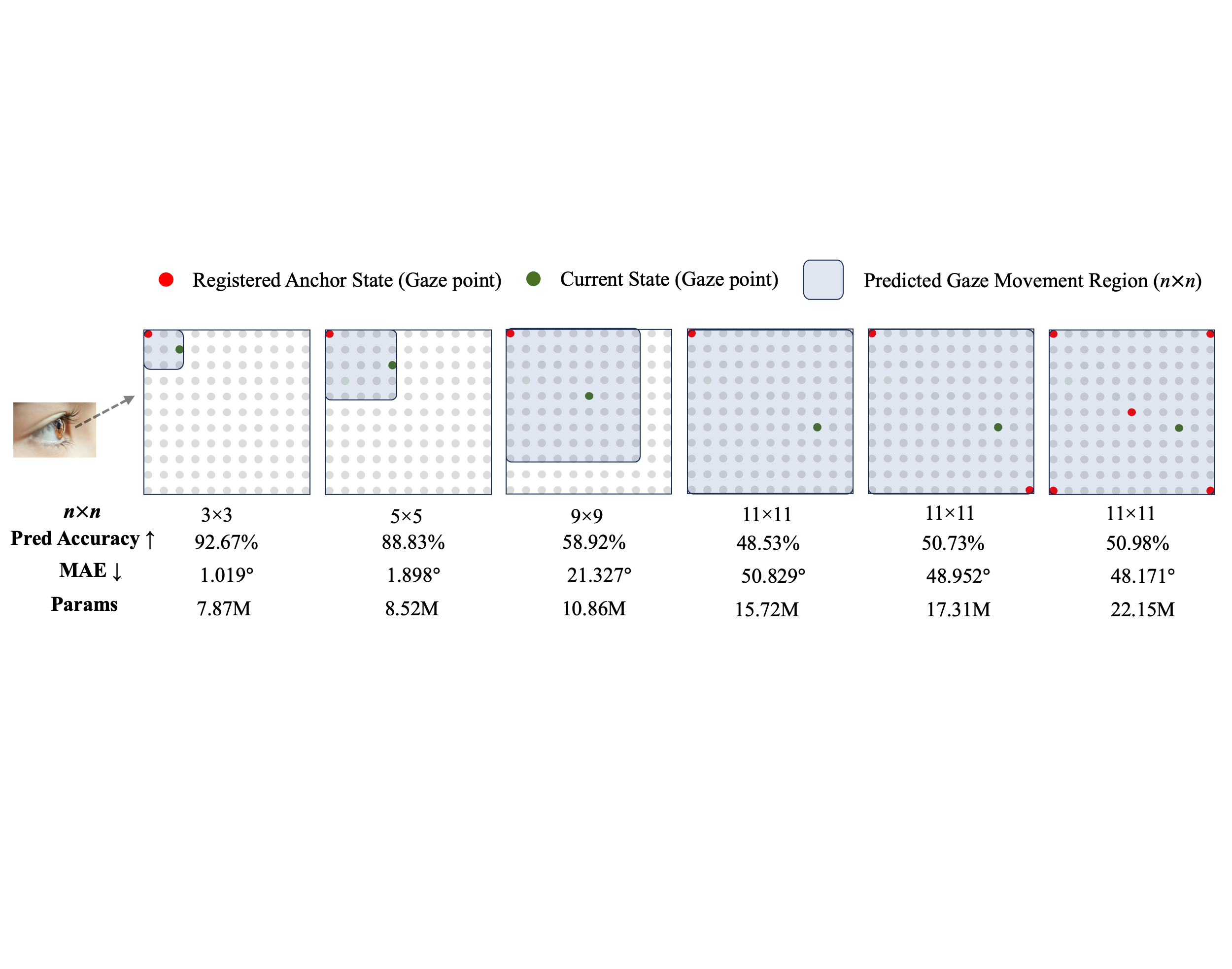}}
    \caption{Illustration of gaze estimation accuracy by trained using different perceived sizes, denoted as $n\times n$. Moreover, all models are evaluated on data with perceived regions identical to those in their respective training sets. The experimental results indicate that an incremental increase in the training dataset region leads to a substantial degradation in network performance. Moreover, the incremental of the network's parameters is for fitting the dataset (otherwise, the network is hard to converge). Meanwhile, as shown in the rightmost example, directly training with multiple anchors in the 11$\times$11 region is also hard to converging on an accurate result. This observation suggests that instead of directly training on the whole region, we can distil those small but accurate models in local regions into a large student network for accurate modelling of gaze motion. $\uparrow$ (resp. $\downarrow$) indicates that larger (resp. smaller) values are better.} 
    \label{fig:comparison}
\end{figure*}

Specifically, due to the different individual biometric characteristics, e.g., pupil distance and size of the eyeball, there would be a significant bias in the gaze estimation process~\cite{Duchowski2017eye}. Consequently, instead of directly calculating the \textbf{absolute location} of gaze focus from a single observational state, we propose to calculate the \textbf{relative shift} of the measured state compared with pre-registered anchor states. Moreover, to learn the correlation between those two states in an end-to-end manner, we delve into the potential of utilizing pre-trained vision transformers for cross-modal eye tracking. However, directly training the gaze estimation network in a large region usually leads to overfitting phenomena, as shown in Fig.~\ref{fig:comparison}. Thus, we train a set of sub-region gaze estimation models for different anchor states, which are called local expert networks (see Sec.~\ref{Sec:Gaze_struct}). 

Subsequently, to further boost the capacity for accurate gaze prediction across diverse scenarios, we design a distillation-based algorithm to ensemble knowledge of pre-trained local expert networks into a large student network. However, the presence of noise in the measured inputs (especially for the event data) can disrupt neural network training and negatively impact performance. To alleviate the potential noise influencing the neural network training, we propose a self-supervised latent denoising neural network for feature maps from experts and then apply knowledge distillation to the student network (see Sec.~\ref{Sec:Distillation}). 

In what follows, we will detail the proposed pipeline.

\begin{figure*}[t]
    \centering
    \resizebox{1\textwidth}{!}{\includegraphics[trim={0cm 0.4cm 0cm 0cm},clip]{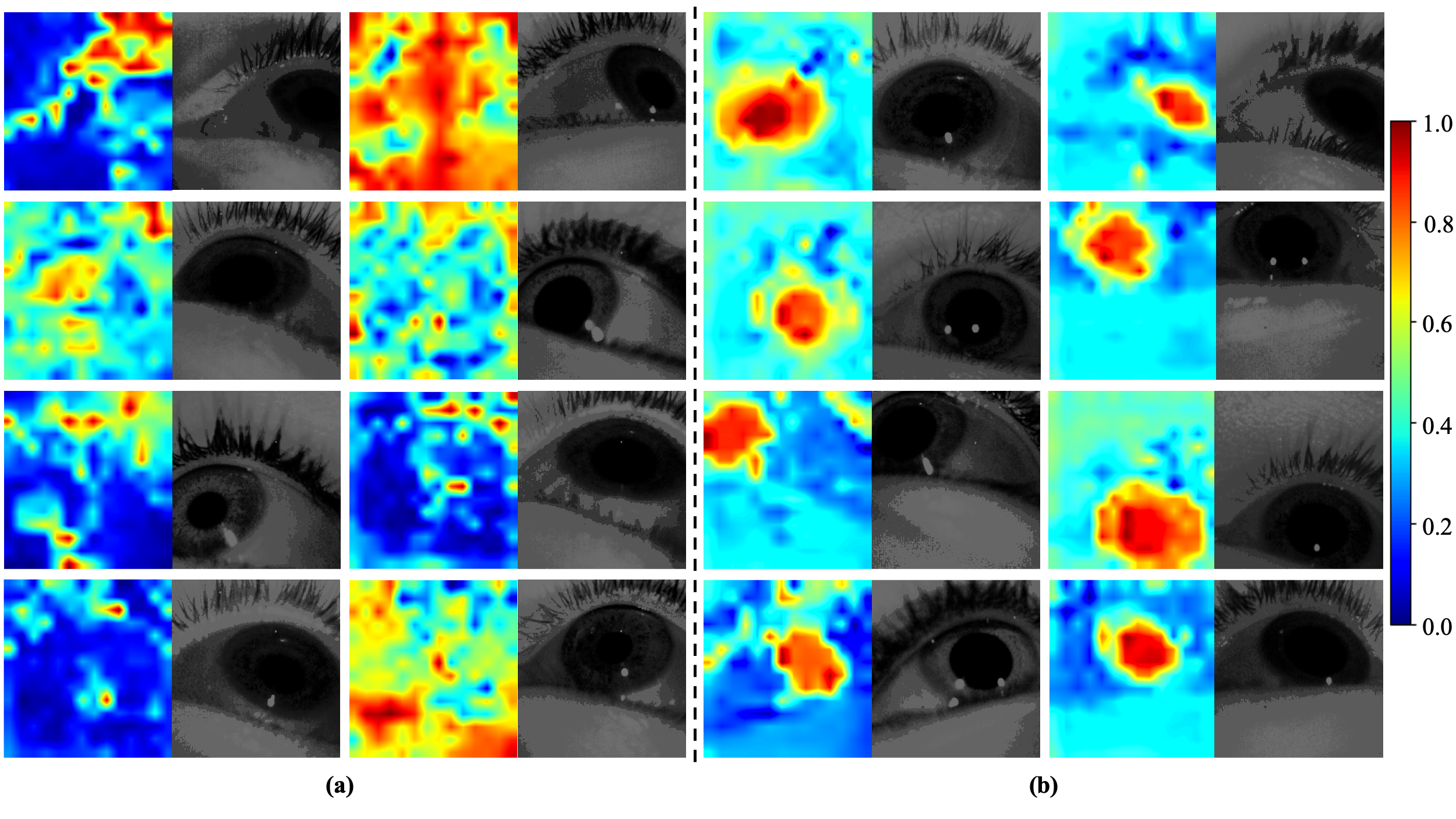}}
    \caption{
    \textbf{(a)} illustrates the outcome of training with a model on a large region, exhibiting pronounced over-fitting, as evidenced by the heatmap, indicating attention dispersion away from the ocular region. \textbf{(b)} showcases the performance of our model, which makes distillation of knowledge from a set of local experts, with a heatmap that is distinctly concentrated on the ocular region. The visualization indicates that through the proposed local-global distillation, network has accurate attention on the relevant region.}\vspace{-0.4cm}
    \label{fig:overfitting}
\end{figure*}

\subsection{State Correlation Modeling by Local Expert}
\label{Sec:Gaze_struct}

\noindent \textbf{Event-Frame Tokenization.} 
Our framework takes a paired near-eye frame and the corresponding event stream as input. To effectively fuse these two data modalities, we transform them into a unified representation. Specifically, we first aggregate events within specific time intervals to convert the asynchronous event flow into a synchronous format, aligning with frame exposure duration. We then voxelize the original event stream into a grid of voxels using PointNet~\cite{charles2017pointnet}. This voxelization enables us to represent the event data in a structured format analogous to the frame data. Finally, we tokenize both the frame and the voxelized event data into sequences of spatial tokens, where each token corresponds to a specific location in the frame and its associated event voxel. This tokenized representation facilitates seamless fusion and processing of event and frame information within our model.

\vspace{+0.2cm}
\noindent \textbf{Correlation Modeling.} 
After creating the unified representation for both data modalities,
we construct the embedding representations of two gaze points, i.e., the estimation gaze point indicates the current state and the template gaze point represents the registered anchor state, by concatenating the token embeddings of two modalities from the same state.

We then apply a Vision Transformer (ViT) with the multi-head self-attention (MSA) mechanism to model the correlation between the current and registered anchor states. The processed features are subsequently flattened into a single feature representation, which is analyzed by a convolutional layer to predict the final class label.

\begin{figure}[t]
    \centering
    \resizebox{0.5\textwidth}{!}{\includegraphics[trim={0cm 0.4cm 0cm 0cm},clip]{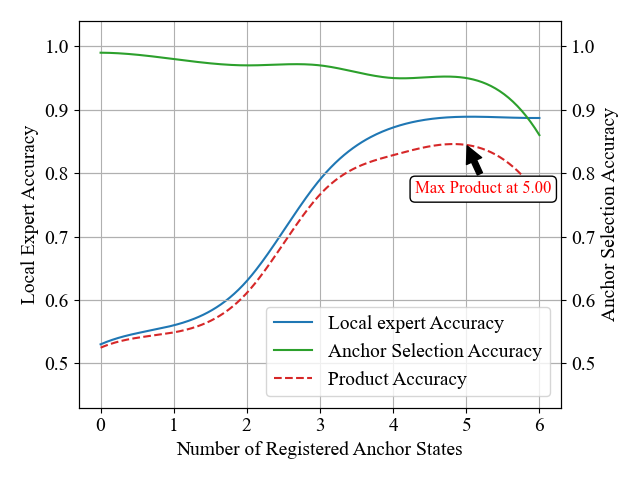}}
    \caption{Impact of the number of registered anchor states on prediction accuracy. The results demonstrate that increasing the number of anchor states generally improves prediction accuracy, with an optimal performance achieved when using 5 anchor states. \textit{Note that product accuracy can be observed on both vertical axes.}
    }\vspace{-0.4cm}
    \label{fig:anchorsetting}
\end{figure}

Our experiments confirm the model's effectiveness in distinguishing between the current and registered anchor states for gaze estimation. However, we observed limitations when a single registered anchor state was used for predictions across a large gaze area, leading to significant over-fitting as shown in Figs.~\ref{fig:comparison} and \ref{fig:overfitting}.
To address this, we propose partitioning the entire gaze point area into $N$ smaller sub-regions, each with its own regional registered anchor state. This allows us to train $N$ specialized local expert networks. For our task, we achieve a good balance between training costs and accuracy by dividing the region into five sub-regions, as shown in Fig.~\ref{fig:anchorsetting}, each with a dedicated local expert network.

\begin{figure*}[t]
    \centering
    \resizebox{1\textwidth}{!}{\includegraphics[trim={2.5cm 9cm 2.5cm 3.5cm},clip]{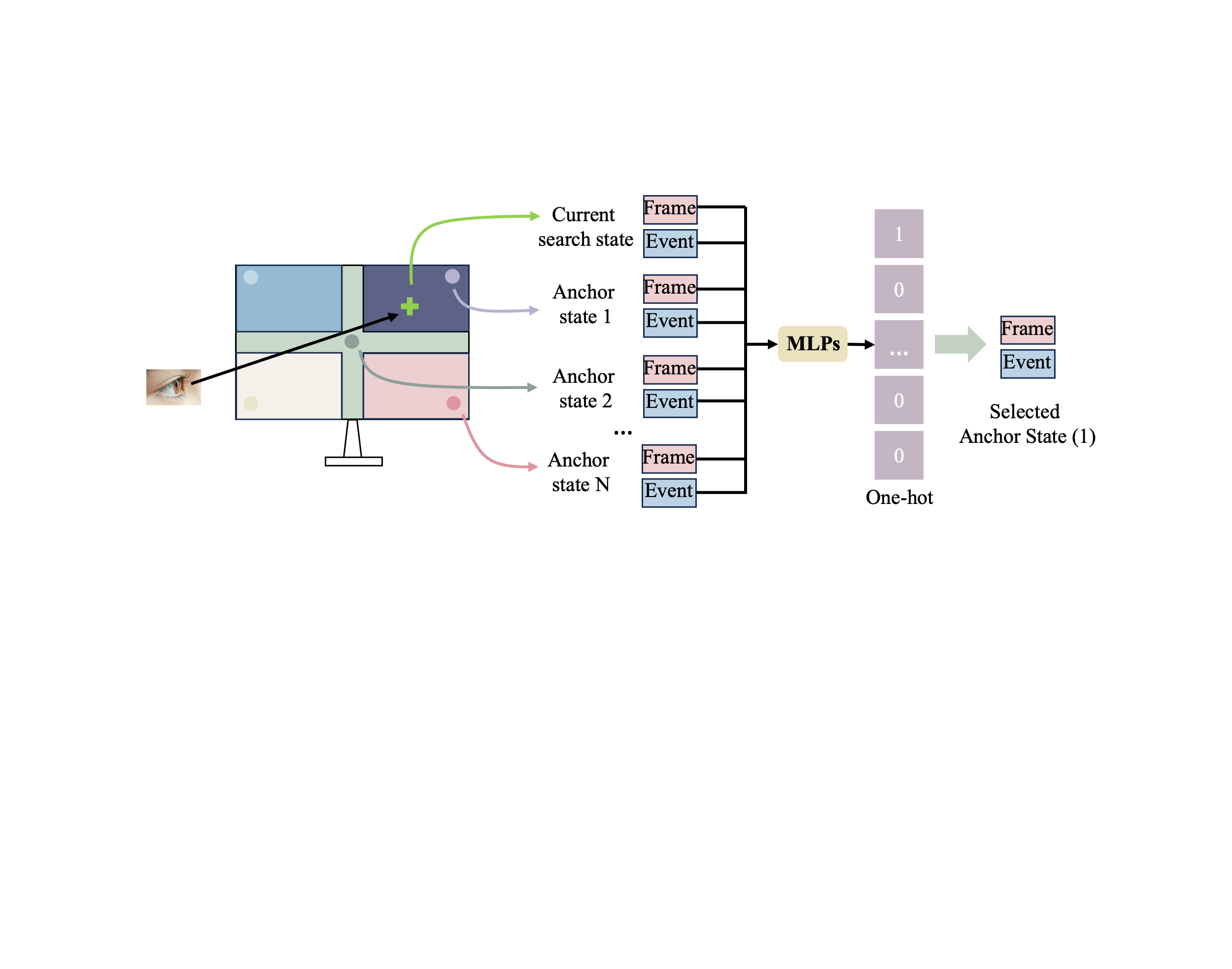}} \vspace{-0.5cm}
    \caption{Illustration of the \textbf{anchor selection} mechanism. we employ an MLP-driven anchor selection mechanism to dynamically identify the nearest anchor state for the input state. The correlation with other modules is illustrated in the training part in Fig.~\ref{fig:framework}.}
    \label{fig:anchorselection}
\end{figure*}

\vspace{0.2cm}
\noindent \textbf{Anchor Selection Mechanism.} Since the local expert networks only perceive data from their own sub-regions, we further design a network to globally determine the usage of which expert (anchor) during the gaze estimation process. As illustrated in Fig.~\ref{fig:anchorselection}, the network uses the MLPs to choose the nearest registered anchor state based on the input current state to determine which sub-region the current state belongs to.

\subsection{Local-Global Latent Denoising Distillation}
\label{Sec:Distillation}

\begin{figure}[t]
  \centering
  \begin{minipage}[t]{1\linewidth}
    \begin{algorithm}[H]
      \caption{Training Denoiser}
      \label{alg1}
      \begin{algorithmic}[1]
        \State \textbf{Repeat}
        \State $\mathbf{X}_i \sim q(\mathbf{X}_i)$
        \State $t \sim \mathrm{Uniform}(\{i,i+1,...,T\})$
        \State $\epsilon \sim\mathcal{N}(0, \mathbf{I})$
        \State $i \sim \mathcal{U}(27, 32)$
        \State Take gradient descent step on
        \State ~~~~ $\delta = \epsilon-\epsilon_\theta(\sqrt{\frac{\Bar{\alpha}_t}{\Bar{\alpha}_i}}\mathbf{X}_i+\sqrt{1 - \frac{\Bar{\alpha}_t}{\Bar{\alpha}_i}}\epsilon,t)$
        \State ~~~~$\nabla_\theta \| \sum \delta \|^2 + \| \sigma(\delta) - \sqrt{2} \|^2$ 
        \label{alg1:closedform}
        \State \textbf{Until} converged
      \end{algorithmic}
    \end{algorithm}
  \end{minipage}
  \hfill
  \begin{minipage}[t]{1\linewidth}
    \begin{algorithm}[H]
    \label{line:reverse}
      \caption{Reverse process}
      \label{alg2}
      \begin{algorithmic}[1]
        \State $\epsilon \sim \mathcal{N}(0, \mathbf{I})$
        \State $\mathbf{X}_{T'}  = \sqrt{\frac{\Bar{\alpha}_{T'}}{\Bar{\alpha}_i}}\mathbf{X}_i+\sqrt{1 - \frac{\Bar{\alpha}_{T'}}{\Bar{\alpha}_i}}\epsilon$
        \State \textbf{For} $t=T',...,i$ \textbf{do}
        \State ~~~~$\mathbf{z} \sim\mathcal{N}(0, \mathbf{I})$ if $t>i$, else $\mathbf{z}=0$
        \State ~~~~$\sigma_t =  \sqrt{\frac{\hat{\beta}_t(\Bar{\alpha}_i - \Bar{\alpha}_{t-1})}{\Bar{\alpha}_i  - \Bar{\alpha}_t}} $
        \State ~~~~$\mathbf{X}_{t-1} = \frac{1}{\sqrt{\alpha_t}}(\mathbf{X}_t - \frac{(1-\alpha_t)\sqrt{\Bar{\alpha}_i}}{\sqrt{\Bar{\alpha}_i-\Bar{\alpha}_t}} \mathbf{\epsilon}_{\theta}( \mathbf{X}_t, t)) + \sigma_t \mathbf{z}$
        \State \textbf{End for}
        \State \textbf{Return} $\mathbf{X}_i$
      \end{algorithmic}
    \end{algorithm}
  \end{minipage}
\end{figure}

The proposed denoising diffusion algorithm consists of the following two steps: 
\begin{enumerate}
    \item  training a denoising diffusion network for removing potential noise as illustrated by Eq.~(\ref{Eq:Diffusion}) and Algorithm~\ref{alg1}; and
    \item  distilling the knowledge from local expert networks into a student network via Eq.~(\ref{Eq:loss}) with the denoising process as Algorithm~\ref{alg2}.
\end{enumerate}

Given the feature maps of measurement $\widetilde{\mathbf{X}} = \mathbf{X} + \gamma \mathbf{\epsilon}$, where $\mathbf{X}$ indicates the expectation and $\mathbf{\epsilon} \sim \mathcal{N}(\mathbf{0}, \mathbf{I})$ represents the random Gaussian noise, we aim to recover the distribution of $\widetilde{\mathbf{X}} \sim \mathcal{N}(\mathbf{X}, \gamma \mathbf{I})$ from a given measurement $\widetilde{\mathbf{X}}$. Specifically, 
we train a denoising diffusion network that can generate latent $\mathbf{\mu}$ from Gaussian noise. Then, by progressively removing the noise from the latent of a given noised measurement, we can finally derive the expectation by averaging several denoising results.

We formulate the distribution of the latent feature maps as $p(\widetilde{\mathbf{X}}|\mathbf{X}) = \mathcal{N}(\widetilde{\mathbf{X}}| \mathbf{X}, \gamma\mathbf{I})$, where $\widetilde{\mathbf{X}}$ indicates the measured noisy latent, $\mathbf{X}$ denotes the corresponding noise-free expectation, and $\gamma\mathbf{I}$ represents the covariance matrix with the independent assumption. Since we only know the noised sample, we further re-parameterize the $\gamma'= 1- \gamma $ and $\mathbf{X} = \sqrt{\gamma'} \mathbf{Y}$.

Subsequently, we have $p(\widetilde{\mathbf{X}}|\mathbf{Y}) = \mathcal{N}(\widetilde{\mathbf{X}}| \sqrt{\gamma'} \mathbf{Y}, \gamma \mathbf{I})$. Recalling the intermediate sample of DDPM~\cite{ho2020denoising}, $\mathbf{X}_{t} = \sqrt{\Bar{\alpha}_i} \mathbf{X}_0 + \sqrt{1-\Bar{\alpha}_i} \mathbf{X}_T$, where $\Bar{\alpha}_i = \prod_{j=0}^i\alpha_i$. Thus, the noised latent $\widetilde{\mathbf{X}}$ could be approximated by a certain step result $\mathbf{X}_i$ in a diffusion reverse process~($\mathbf{X}_T \rightarrow \mathbf{X}_0$), where $\mathbf{X}_0 = \mathbf{Y}$ and $\Bar{\alpha_i} = \gamma'$. We then introduce a process to drive the distribution of $\mathbf{X}_i$ from a sample $\widetilde{\mathbf{X}}$ via diffusion models.

The forward process can be simply constructed by adding noise following $\mathbf{X}_{t} = \sqrt{\Bar{\alpha}_t} \mathbf{X}_{t-1} + \sqrt{1-\Bar{\alpha}_t} \epsilon_t$, where $\epsilon_t$ indicates a random noise. Thus, we can train the denoiser as Algorithm~\ref{alg1} to learn the potential noise from the noised sample. 

In Algorithm~\ref{alg1}, we optimize the diffusion denoiser with the following loss term
\begin{equation}
\label{Eq:Diffusion}
L_{Diff} = \nabla_\theta \| \sum \delta \|^2 + \| \sigma(\delta) - \sqrt{2} \|^2.
\end{equation}
Since $\epsilon \sim \mathcal{N}(\mathbf{0},\mathbf{I})$, we expect  $\epsilon_\theta(\sqrt{\frac{\Bar{\alpha}_t}{\Bar{\alpha}_i}}\mathbf{X}_i+\sqrt{1 - \frac{\Bar{\alpha}_t}{\Bar{\alpha}_i}}\epsilon,t)\sim \mathcal{N}(\mathbf{0},\mathbf{I})$. Thus, we have
\begin{equation}
    \mathbf{\delta} = \mathbf{\epsilon}-\mathbf{\epsilon}_\theta(\sqrt{\frac{\Bar{\alpha}_t}{\Bar{\alpha}_i}}\mathbf{X}_i+\sqrt{1 - \frac{\Bar{\alpha}_t}{\Bar{\alpha}_i}}\mathbf{\epsilon},t), \mathbf{\delta} \sim \mathcal{N}(\mathbf{0},2\mathbf{I}),
\end{equation}
Then we regularize $\sum \delta \rightarrow 0$ and $ \sigma(\delta) \rightarrow \sqrt{2}$, which is exact the training objective Eq.~\eqref{Eq:Diffusion}.

However, our algorithm starts from a noised measurement and iteratively adds and then removes noise. Thus, \textbf{the reverse process of our algorithm is explicitly different from the standard de-noising diffusion model}. We further delicately investigate the reverse distribution of the Markov chain, which starts from $\mathbf{X}_i$. We begin with one reverse transition process as  
\begin{equation}
    q(\mathbf{X}_{t-1}|\mathbf{X}_{t}, \mathbf{X}_i) = \frac{q(\mathbf{X}_t|\mathbf{X}_{t-1}, \mathbf{X}_i)q(\mathbf{X}_{t-1}|\mathbf{X}_i)}{q(\mathbf{X}_{t}| \mathbf{X}_i)},
\end{equation}
\noindent where $\mathbf{X}_t$ indicates the noised sample with inherent noise from collected event data.
Then, we take an investigation of the term  $q(\mathbf{X}_{t-1}|\mathbf{X}_t, \mathbf{X}_i)$.
\begin{align}
&q(\mathbf{X}_{t-1}|\mathbf{X}_t, \mathbf{X}_i) = \frac{q(\mathbf{X}_t|\mathbf{X}_{t-1}, \mathbf{X}_i) q(\mathbf{X}_{t-1}| \mathbf{X}_i)}{q(\mathbf{X}_t| \mathbf{X}_i)}\\ 
                    &= \frac{\mathcal{N}(\mathbf{X}_t, \sqrt{\alpha_t}\mathbf{X}_{t-1}, \sqrt{1-\alpha_t}I) \mathcal{N}(\mathbf{X}_{t-1}, \sqrt{\frac{\Bar{\alpha}_{t-1}}{\Bar{\alpha_i}}}\mathbf{X}_{i}, \sqrt{1 - \frac{\Bar{\alpha}_{t-1}}{ \Bar{\alpha_i}}}I)}{ \mathcal{N}(\mathbf{X}_t, \sqrt{\frac{\Bar{\alpha}_{t}}{\Bar{\alpha_i}}}\mathbf{X}_{i}, \sqrt{1 - \frac{\Bar{\alpha}_{t}}{\Bar{\alpha_i}}}I)}\nonumber \\
                    &\propto \mathcal{N} (\mathbf{X}_{t-1}, 
                    \frac{(1-\alpha_t)(\Bar{\alpha}_i - \Bar{\alpha}_{t-1})}{ \Bar{\alpha}_i - \Bar{\alpha}_t}(\frac{\sqrt{\alpha_t}}{1-\alpha_t} \mathbf{X}_t + \frac{\sqrt{\Bar{\alpha}_{i}\Bar{\alpha}_{t-1}}}{\Bar{\alpha}_{i} - \Bar{\alpha}_{t-1}} \mathbf{X}_{i}), \\
                    &\sqrt{\frac{(1-\alpha_t)(\Bar{\alpha}_i - \Bar{\alpha}_t)}{\Bar{\alpha}_i  - \Bar{\alpha}_t}} ). \nonumber
\end{align}

\noindent Moreover, with $\hat{\beta}_t = 1 - \alpha_t$, we have 
\begin{equation}
\begin{aligned}
\label{Eq:Noise}
q(\mathbf{X}_{t-1}|\mathbf{X}_t, \mathbf{X}_i) \propto \mathcal{N}(\mathbf{X}_{t-1}, 
                    \frac{\hat{\beta}_t(\Bar{\alpha}_i  - \Bar{\alpha}_{t-1})}{ \Bar{\alpha}_i - \Bar{\alpha}_t}(\frac{\sqrt{\alpha_t}}{\hat{\beta}_t} \mathbf{X}_t + \\
                    \frac{\sqrt{\Bar{\alpha}_i\Bar{\alpha}_{t-1}}}{\Bar{\alpha}_i - \Bar{\alpha}_{t-1}} \mathbf{X}_{i}), \sqrt{\frac{ \hat{\beta}_t(\Bar{\alpha}_i - \Bar{\alpha}_{t-1})}{\Bar{\alpha}_i  - \Bar{\alpha}_t}} ).
\end{aligned}
\end{equation}

However, such a noise level indicated by $i$ is quite hard to derive. 
Thus, during the training phase, we relax $\Bar{\alpha}_i$ to a certain range of $i\in[27,32]$. Note that through the aforementioned reverse process, we derive the distribution of $\mathbf{X}_i$ from its certain measurement by iteratively sample $\mathbf{X}_{t-1}, t={T',~\cdots,~i},$ as Eq.~\eqref{Eq:Noise}, which is detailedly illustrated in Algorithm~\ref{alg2}. Although we do not explicitly remove the noise from the data, we expect during the training process, the variable $\mathbf{X}$ would be converged to the noise-free expectation $\mathbf{X}_0$ as

\begin{align}
\label{Eq:loss}
   &\mathcal{L}_{d}(\mathbf{X}^\theta,\mathbf{X}^\phi_i) = \sum_{\mathbf{X}^\phi_i\sim p(\mathbf{X}^\phi_i)} \| \mathbf{X}^\theta - \frac{\mathbf{X}^\phi_i}{\sqrt{\Bar{\alpha}_i}} \|_2^2,
\end{align}
where $\sum_{\mathbf{X}^\phi_i\sim p(\mathbf{X}_i)}$ indicates summation across different diffusion reconstruction samples of teacher network $\phi$, $\mathbf{X}_\theta$ denotes the output feature of student network $\theta$.

\subsection{Training Objective}
We train our gaze estimation framework in a two-stage manner. In the first stage, we train five local experts, each specializing in a sub-region of the gaze point area.
Each expert is trained using the cross-entropy loss function, $\mathcal{L}_e$, with corresponding subsets of gaze point labels and regional registered anchor states:
\begin{equation}
     \mathcal{L}_e = \mathcal{L}_{CE}(\hat{\mathbf{Y}},\mathbf{Y}),
\end{equation}
where $\hat{\mathbf{Y}}\in\mathbb{R}^{L}$ and $\mathbf{Y}\in\mathbb{R}^{L}$ represent the predicted and ground-truth locations, respectively.

The second stage involves distilling the knowledge from these local experts into a student network designed to handle the entire gaze movement region. 
This distillation process utilizes three loss functions. We maintain the use of $\mathcal{L}_e$ as a hard loss to ensure stable performance of the student network. 
Additionally, we employ a soft loss, the Kullback-Leibler Divergence loss function $\mathcal{L}_s$ to guide the learning of the student network:
\begin{equation}
     \mathcal{L}_s = \mathcal{L}_{KL}(\mathbf{T}_{S}, \mathbf{T}_{E}),
\end{equation}
where $\mathbf{T}_{S}, \mathbf{T}_{E}\in \mathbb{R}^{H\times W}$ represent the attention matrices of the student and local experts, respectively. Finally, the total loss for the second stage is a weighted sum of these three losses:
\begin{equation}
     \mathcal{L} = \alpha \cdot \mathcal{L}_e + \beta \cdot \mathcal{L}_s + \lambda \cdot \mathcal{L}_{d},
\end{equation}
where $\alpha$, $\beta$, and $\lambda$ are corresponding weights for balancing different loss terms. Based on our extensive ablation experiments, we set $\alpha$, $\beta$, and $\lambda$ to 1, 1, and 500, respectively.

\section{Experiment}
\label{sec:experiment}
\subsection{Experimental Settings}
\mypara{Dataset} 
We utilized a hybrid event-based IR near-eye gaze tracking dataset \cite{Angelopoulos2021event} to evaluate our proposed system. The dataset integrated a sophisticated DAVIS-346b sensor (iniVation) with a high-resolution 25 mm f/1.4 VIS-NIR C-mount lens (EO-\#67-715), further augmented with a UV/VIS cutoff filter (EO-\#89-834) to capture the ocular dynamics of subjects using an ophthalmic headrest coupled with a restraining apparatus to minimize potential head movement artifacts. 

The dataset recorded the gaze movements from 24 subjects using a 40$\times$40-pixel luminous green fixation cross rendered on a 40-inch-diagonal, 1920 $\times$ 1080-pixel display (Sceptre 1080p X415BV-FSR) with a visual field of view (FoV) of 64$^\circ \times$ 96$^\circ$. 

\begin{figure*}[t]
    \centering
    \resizebox{1\textwidth}{!}{\includegraphics[trim={0cm 6.0cm 0cm 5.5cm},clip]{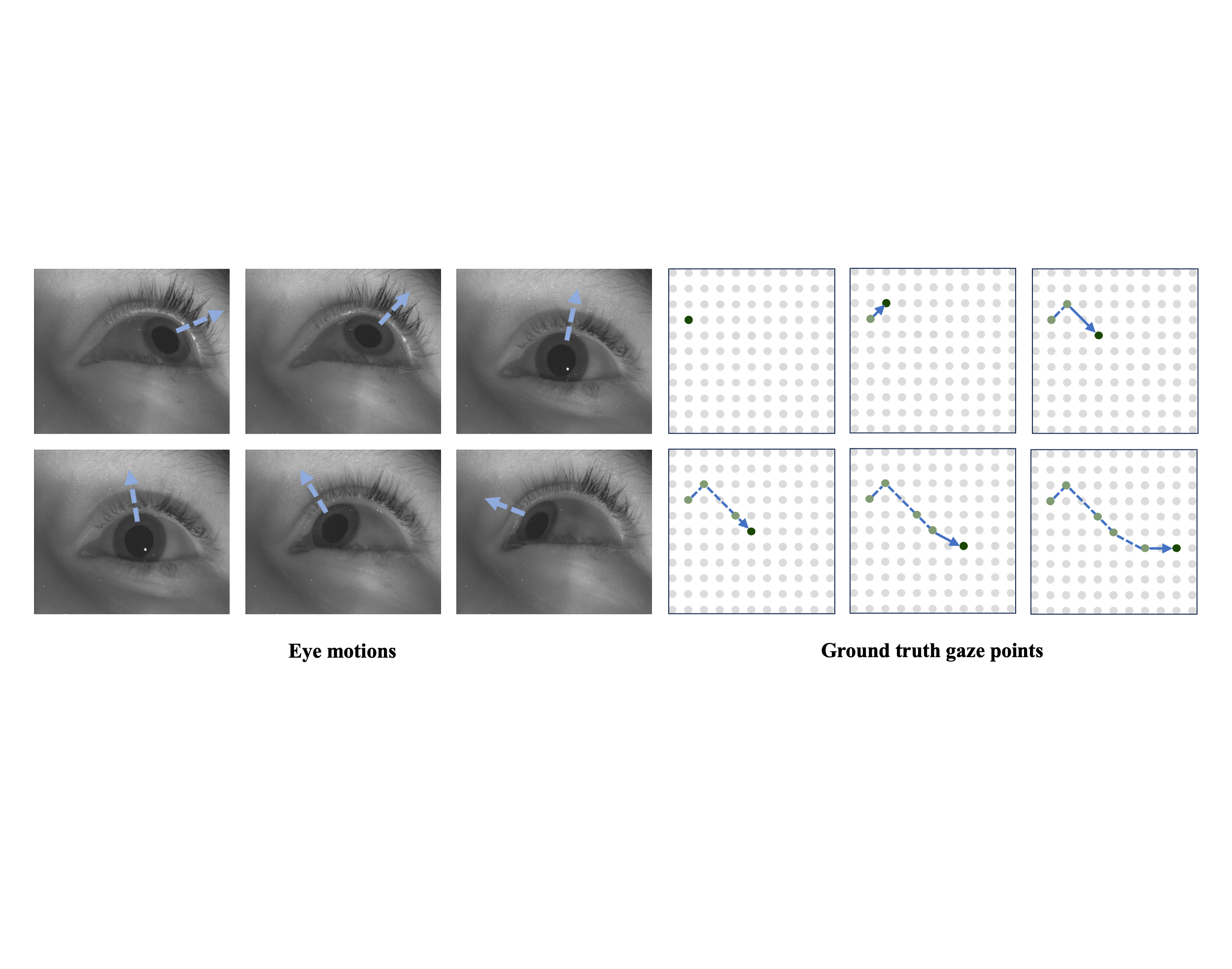}}
    \caption{Illustrative depictions of ocular motion trajectories and the associated gaze coordinates on the visual field.
    }
    \label{fig:eye_motion}
\end{figure*}

The dataset was divided into two distinct experimental conditions tailored to elicit different oculomotor responses: \textit{stochastic saccadic movements} and \textit{controlled smooth pursuit tracking}. During the first experimental paradigm, subjects were instructed to direct their gaze towards the stimulus. The stimulus materialized at random within a grid matrix of 121 discrete points (an 11$\times$11 grid pattern projected onto the display), with each point being presented for 1.5 seconds. This sequence of locations was uniformly randomized and remained consistent across all participants. Some sample images are shown in Fig. \ref{fig:eye_motion}. The second experimental paradigm was used for free gaze point estimation. In this paradigm, the subjects' task was to maintain visual fixation on the stimulus as it traversed a predefined square-wave trajectory. This trajectory commenced at the upper boundary of the display and proceeded in a downward motion, covering the full horizontal extent of the screen, with a vertical displacement amplitude of 150 pixels. Despite the intentional induction of saccadic jumps and smooth pursuit movements within the experimental framework, the resulting dataset encapsulates a plethora of involuntary eye dynamics, including microsaccades and ocular tremors, thereby offering a comprehensive profile of ocular motion behavior.

\vspace{+0.2cm}
\mypara{Implementation details} \noindent \textbf{1) Local expert network.} We set the input frame and event
voxel size to 224$\times$224. We directly adopt the first two layers of the base vision transformer (ViT-B). We configure the mini-batch size to 80 and adopt the AdamW optimization algorithm due to its efficacy in handling sparse gradients and incorporating weight decay for regularization. We initiate the training with a learning rate of 0.0001, which is methodically attenuated following a cosine annealing schedule, descending to a factor of 0.1 of the original learning rate. The training epochs are set to 350 to ensure the local expert networks adequately learn the intricate patterns within the data. \noindent \textbf{2) Distilled student network.} In the subsequent stage, the five local expert networks are distilled into the student network, leveraging the AdamW optimizer once again. The student network consists of the first two layers of ViT-B. This phase is conducted with a reduced learning rate of $10^{-5}$ and a momentum coefficient of 0.9, extending over 500 epochs. Due to the computationally intensive nature of knowledge distillation, the mini-batch size is adjusted to 4 to accommodate the increased model complexity and ensure stable convergence. Code is implemented using Pytorch and trained with RTX3090 GPUs.

\vspace{+0.2cm}
\mypara{Inference} The inference of our model is different from the training process. During training, it is necessary to do latent denoising, to convert the feature maps into a distribution before distilling its knowledge into the student model. During inference, it only requires the use of the MLPs to determine which sub-region the current state belongs to and directly feeds the fused input into the student network without doing latent denoising, as the student model already possesses complete inference capabilities, as shown by the pink arrow in Fig.~\ref{fig:framework}

\vspace{0.2cm}
\noindent\textbf{Metrics.} Following previous works \cite{chen2022real}, we employed the Mean Angle Error (MAE) to evaluate the performance quantitatively, computed as
\begin{equation}
    \mathrm{MAE} = \frac{1}{N}\sum_{i=1}^N \arccos{\frac{<\overrightarrow{p}_i, \overrightarrow{t}_i>}{\| \overrightarrow{p}_i\| \|\overrightarrow{t}_i\|}}
\end{equation}
where $N$ is the number of samples in the dataset, $\overrightarrow{p_i}$ is the predicted gaze vector for the \textit{i}-th sample, $\overrightarrow{t_i}$ is the corresponding ground truth vector, and $<\cdot, \cdot>$ computes the inner product of two input vectors. 

\subsection{Comparison with State-of-the-Art Methods}
 We compared our method with the following four methods:
\begin{itemize}
    \item \textbf{S-T GE} \cite{Palmero2020benefits} leverages temporal sequences of eye images to enhance the accuracy of an end-to-end appearance-based deep-learning model for gaze estimation.

    \item  \textbf{Dilated-Net} \cite{chen2019appearance}~integrates dilated convolutional layers to enhance feature extraction for gaze estimation.
    \item \textbf{EventGT} \cite{Angelopoulos2021event} achieves an equilibrium of computational efficiency and accuracy—a key consideration for real-time gaze tracking.

    \item \textbf{HE-Tracker} \cite{chen2022real} employs a sophisticated pipeline that commences with the E-Tracker's encoding of eye imagery.
\end{itemize}

The quantitative results are presented in Table \ref{table:baseline}, where it can be seen that our method achieves superior performance via reducing mean absolute error by nearly $50\%$ and boosting tracking accuracy by $15\%$ compared to recent state-of-the-art methods, confirming the effectiveness of our framework. We also refer the reviewers to the \textbf{video demo} for more visual results.

\begin{table*}[t]
	\caption{Quantitative results of different methods. $\downarrow$ (resp. $\uparrow$) indicates the smaller (resp. larger), the better.}
  \vspace{-0.3cm}
 \label{table:baseline}
	\centering
	\footnotesize
	\renewcommand{\arraystretch}{1.5}
	\setlength{\tabcolsep}{5.0mm}
	\begin{tabular}{c | c c c c c }
		\toprule[1.2pt]
		 Method & S-T GE\cite{Palmero2020benefits} & Dilated-Net\cite{chen2019appearance} & EventGT\cite{Angelopoulos2021event} & HE-Tracker\cite{chen2022real} & Ours \\
        \hline 
        MAE $\downarrow$  &7.650$^\circ$  &4.020$^\circ$  &3.000$^\circ$ &4.170$^\circ$	 &\textbf{1.928$^\circ$}  \\
		Accuracy $\uparrow$  &61.88\% &66.63\%  &72.06\% &72.87\%	&\textbf{87.67\%} \\
        Time (ms) &288.34  &562.03 &--- &20.65	 &191.25  \\
	\bottomrule[1.2pt]
	\end{tabular}
    
\end{table*}

\subsection{Ablation Study}

\mypara{Data Modality} 
We analyzed the impact of different data modalities on gaze estimation performance in Table \ref{table:data modality}. Using only frames resulted in a substantial angular error of $40.16^\circ$, significantly higher than the two-modality baseline of $1.93^\circ$. Conversely, relying solely on event data yielded a poor prediction accuracy of $1.45\%$. These findings highlight that although both frames and event data correlate with the directional vector of gaze, their individual use is insufficient for accurate gaze estimation. The best performance is achieved by fusing frames and event data, underscoring the value of our multimodal approach.

\vspace{+0.2cm}
\mypara{Latent Denoising} We also conducted experiments to validate the effectiveness of latent denoising. As shown in Table.~\ref{table:data modality}, incorporating latent denoising enables our method to achieve nearly a $3\%$ improvement in accuracy and a $44.5\%$ reduction in MAE, showcasing the potential of the proposed denoising distillation strategy.

\begin{table}[t]
	\caption{Gaze estimation performance across various data modalities and denoising, where 'F' indicates the frame and 'E' represents the event.}
  \vspace{-0.3cm}
 \label{table:data modality}
	\centering
	\footnotesize
	\renewcommand{\arraystretch}{1.5}
	\setlength{\tabcolsep}{2.0mm}
	\begin{tabular}{c | c c c c }
		\toprule[1.2pt]
		Metric & F & E & Ours~(F+E) & Ours w/o Denoising\\
        \hline 
        MAE $\downarrow$  &40.161$^\circ$  &---  &1.928$^\circ$	 & 3.472$^\circ$\\
        Accuracy $\uparrow$ &53.00\%  &1.45\%  &87.67\%  & 84.64\% \\
	\bottomrule[1.2pt]
	\end{tabular}
\end{table}

\begin{figure}[t]
    \centering
        \resizebox{0.49\textwidth}{!}{\includegraphics[trim={6cm 0.5cm 5cm 1cm},clip]{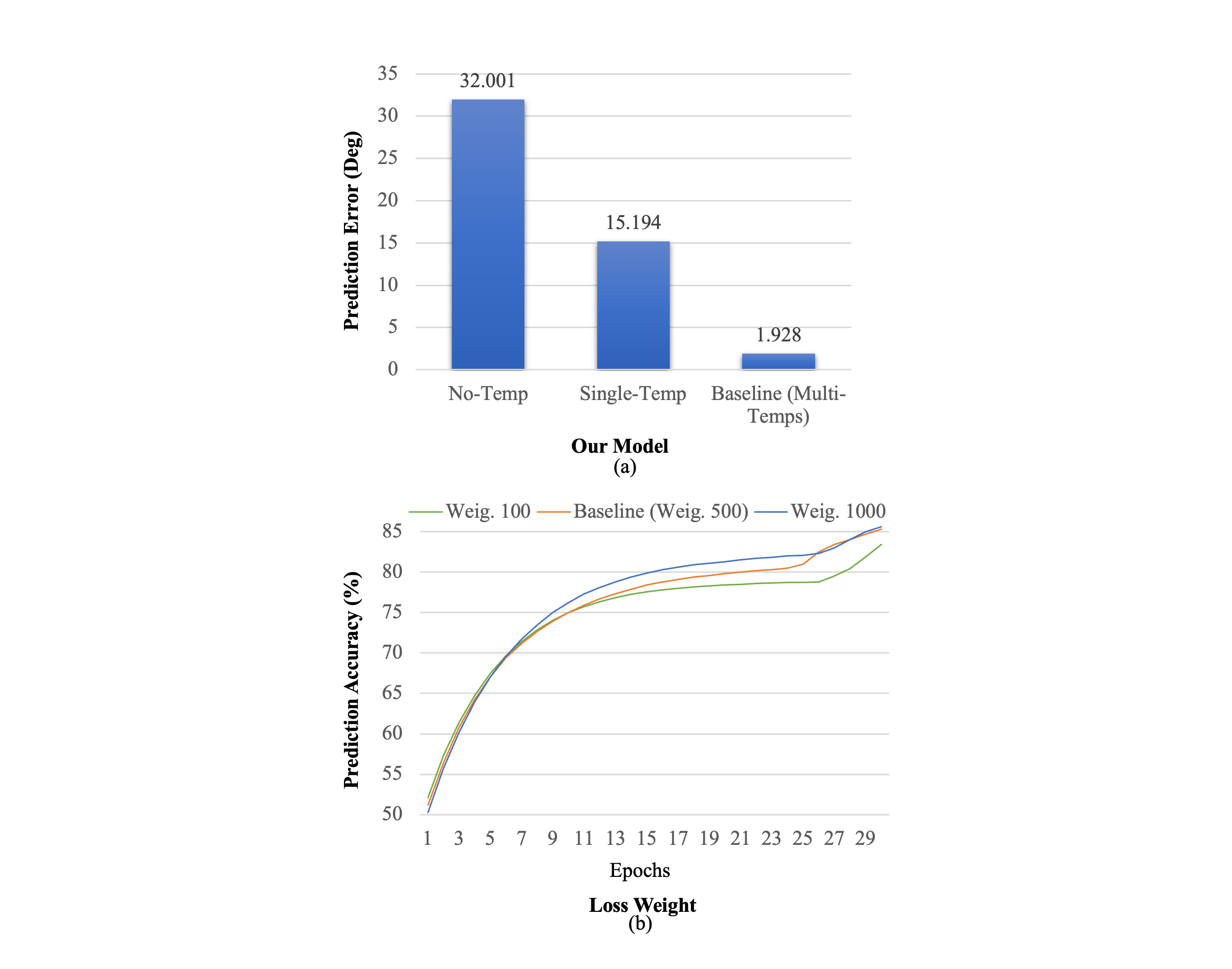}}
        \caption{\textbf{(a)} Performance across different numbers of registered anchor states.  \textbf{(b)} Comparative analysis of model performance subject to varying weights assigned to feature map loss.
    }
    \label{fig:comparison of calib. state and loss weight}
\end{figure}

\vspace{+0.2cm}
\mypara{Anchor State} Meanwhile, we conducted a series of ablation studies to ascertain the necessity of a registered anchor state on the performance of our gaze estimation model. In the absence of any anchor state, our model yielded gaze estimation with an angular error of 32.00$^\circ$. In marked divergence, a single registered anchor state resulted in substantial mitigation of angular error to 15.19$^\circ$, thereby highlighting the importance of the registered anchor state in the enhancement of gaze prediction fidelity. These ablation experiments also confirmed the effectiveness of introducing multiple registered anchor states for tasks involving large gaze movement regions, as mentioned in Sec.~\ref{Sec:Gaze_struct}. 
These empirical findings are concisely encapsulated in Fig.~\ref{fig:comparison of calib. state and loss weight} (\textcolor{red}{a}).

\vspace{+0.2cm}
\noindent \textbf{Weight of Feature Map Loss.} Within our model, a feature map loss $\mathcal{L}_d$  is employed to the guidance of the learning algorithm. 
Note that the gradient from MSE loss (feature distillation) is typically smaller than KL-divergence (task loss), to balance those different terms, we give a large factor for distillation loss $\mathcal{L}_d$. As shown in Fig. \ref{fig:comparison of calib. state and loss weight}~(\textcolor{red}{b}), it can be seen that the weight of distillation loss is quite large. Such an increment is imperative to augment the network's learning capacity, thereby enhancing the precision of gaze estimation.

\vspace{+0.2cm}
\noindent \textbf{Number of Distillation Samples.} As shown in Eq.~\eqref{Eq:loss}, we expect the network can learn the noise-free samples from multiple noised measurements. Thus, enlarging the batch size is a necessary step to make the solution value of Eq.~\eqref{Eq:loss} converge to expectation. We investigated the impact of varying the number of distillation samples (indicating the number of summation samples in Eq.~\ref{Eq:loss}), including 4, 8, and 16 samples.  The results are shown in Table \ref{table:ablation}. The results indicate a positive correlation between the number of distillation samples and the accuracy of gaze estimation. Meanwhile, there is an observable decrease in the MAE. It indicates that without enough samples, it may potentially reduce the model's predictive capabilities, since the gradients of Eq.~\eqref{Eq:loss} in small batches may lead to deteriorated model weight distributions. In other words, the results demonstrate that this approach effectively reduces the over-fitting issue.

\begin{table}[t]
	\caption{Gaze estimation performance across various reconstruction samples. The adopted strategy is underlined. 
 }
 \vspace{-0.3cm}
 \label{table:ablation}
	\centering
	\footnotesize
	\renewcommand{\arraystretch}{1.5}
	\setlength{\tabcolsep}{3.3mm}
	\begin{tabular}{c | c c c  }
		\toprule[1.2pt]
		 Metric & 4 Samples & 8 Samples & \underline{16 Samples} \\
        \hline 
         MAE $\downarrow$&3.987$^\circ$  &2.684$^\circ$  &1.928$^\circ$	  \\
         Accuracy $\uparrow$&82.01\%  &85.48\%  &87.67\% \\
	\bottomrule[1.2pt]
	\end{tabular}
\end{table}

\subsection{Aligning Continuous Location Prediction into Pre-trained Model}
\label{Sec:Continous}
\begin{figure*}[t]
    \centering
    \resizebox{1\textwidth}{!}{\includegraphics[trim={0cm 6.3cm 0.5cm 6cm},clip]{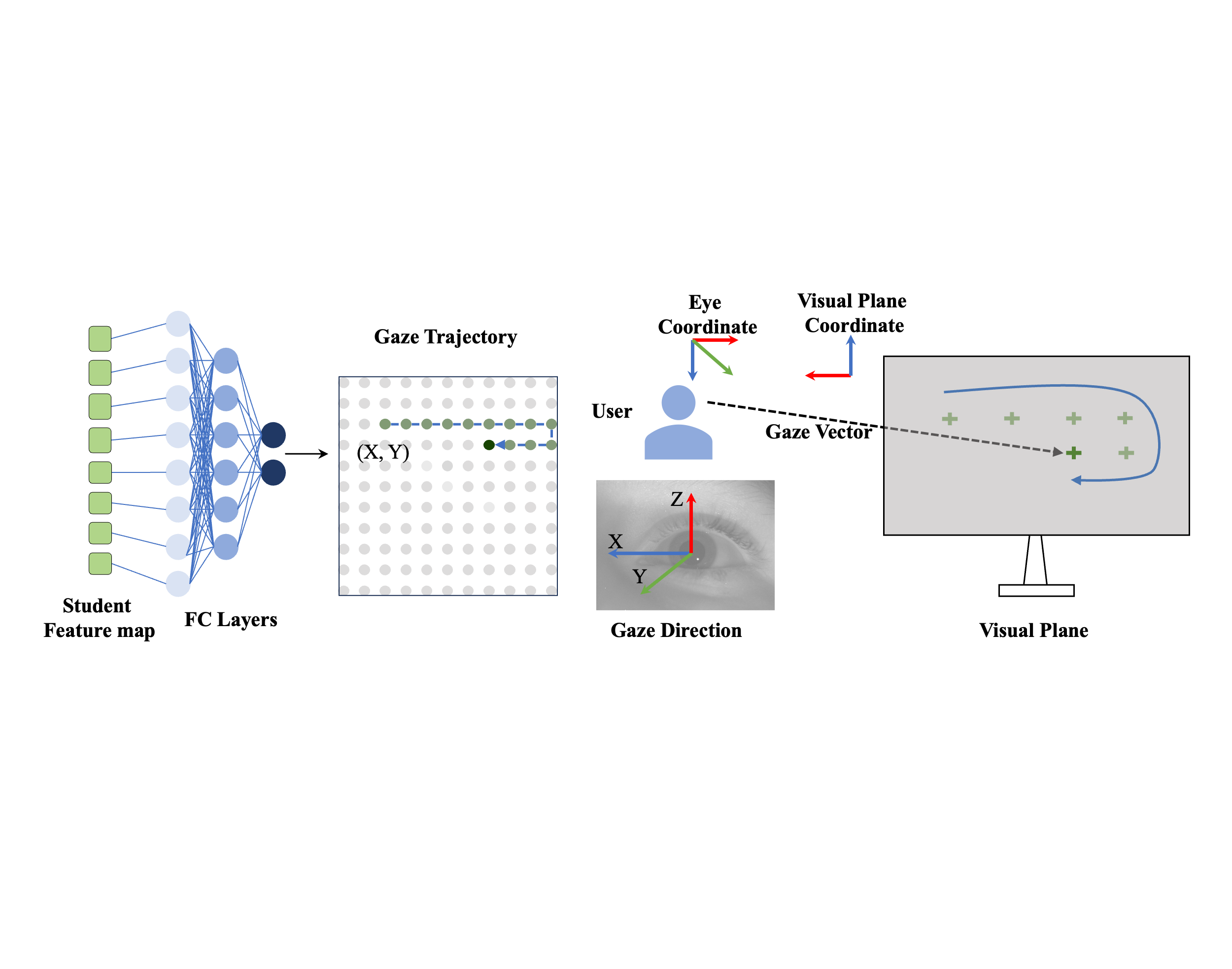}}\vspace{-0cm}
    \caption{Integration of a continuous location prediction network for eye tracking within a pre-trained model via fully connected layer units.
    }\vspace{-0.cm}
    \label{fig:continuous}
\end{figure*}

An essential requirement of eye tracking technology is the capability to dynamically estimate the point of gaze. After distilling local experts into a comprehensive model, our system can provide a rough estimate of the gaze locus with low resolution. Building upon the foundation of a pre-trained model, we generate accurate three-dimensional coordinates for the free gaze point~\cite{Duchowski20223D,wang20183D,Mansouryar20163D}. To produce the actual location of the gaze intersection with the screen, we have designed a branch for gaze projection coordinates, as depicted in Fig.~\ref{fig:continuous}. Specifically, we have adapted the final output layer, traditionally responsible for generating class labels, to directly predict the two-dimensional coordinates of the gaze point on the screen~\cite{Elmadjian20183D,man20133D}. Concurrently, we refined the optimization objective of the model by transitioning to an alternative loss function, designated as $\mathcal{L}_c$, to fine-tune the pre-trained model parameters. The MSE loss function was selected to direct the training process of the gaze position regression model, providing a robust quantitative measure for minimizing the discrepancy between the predicted and actual gaze coordinates. This strategic modification is predicated on enhancing the model's precision in capturing the subtleties of gaze behavior.

\begin{equation}
    \mathcal{L}_c = \| \hat{\mathbf{P}}- \mathbf{P}\|_2^2,
\end{equation}
where $\hat{\mathbf{P}}\in\mathbb{R}^{2}$ and $\mathbf{P}\in\mathbb{R}^{2}$ represent the predicted value of the model and the ground-truth value of the gaze point, respectively. 

This modification enables the system to translate the abstract understanding of where a person is looking into a concrete set of screen coordinates, facilitating applications that require precise tracking of the user's point of gaze. Next, our system establishes a spatial coordinate system with the human eye as the origin in three-dimensional space. Subsequently, using randomly captured near-eye frame and event data, our system outputs continuous location predictions. The quantitative results of this assessment are systematically presented in Table~\ref{table:freepoints}. We also refer reviewers to the \textbf{video demo} for more results.

\begin{table}[t]
	\caption{Quantitative results of our methods of continuous location prediction. $\downarrow$ indicates the smaller, the better.}
  \vspace{-0cm}
 \label{table:freepoints}
	\centering
	\footnotesize
	\renewcommand{\arraystretch}{1.5}
	\setlength{\tabcolsep}{0.5mm}
	\begin{tabular}{c | c c c c c }
		\toprule[1.2pt]
		 Method & S-T GE\cite{Palmero2020benefits} & Dilated-Net\cite{chen2019appearance} & EventGT\cite{Angelopoulos2021event} & HE-Tracker\cite{chen2022real} & Ours \\
        \hline 
        MAE $\downarrow$  &6.980$^\circ$  &3.589$^\circ$  &3.900$^\circ$ &3.655$^\circ$	 &\textbf{3.184$^\circ$}  \\
		
	\bottomrule[1.2pt]
	\end{tabular}
    \vspace{-0.5cm}
\end{table}

\section{Conclusion and Discussion}
\label{sec:conclusion}
We have presented a novel coarse-to-fine dual-stage model for gaze estimation that leverages frame data and event data, utilizing anchor states to enhance precision. Technically, we employed event-frame transformer as backbone and introduced a global-local latent denoising knowledge distillation to effectively merge the unique attributes of frame and event data. Our extensive experiments confirm the model's capability to tackle challenges in multimodal data fusion and reduce overfitting tendencies. Our approach achieves reliable gaze estimation, maintaining angular error below 2$^\circ$. This outperforms various contemporary state-of-the-art gaze estimation methods, setting a new standard for this intricate task.

Despite the substantial superiority of the proposed method compared to state-of-the-art approaches, additional considerations must be addressed for further advancement. Focusing on the development of lightweight neural networks to optimize the inference process is essential. This can be accomplished through techniques like distillation or neural network pruning. Furthermore, there is scope for enhancing accuracy at the retina level (below $1^\circ$), akin to the exceptional capabilities demonstrated by Apple Vision Pro and HTC Vive.

%
\bibliographystyle{IEEEtran}
\bibliography{reference}

\end{document}